# Machine learning-based approach for online fault Diagnosis of Discrete Event System

**R. Saddem, D. Baptiste**

*CReSTIC UFR Exact and Natural Sciences Reims, France*
*ramla.saddem@univ-reims.fr*

**Abstract**: The problem considered in this paper is the online diagnosis of Automated Production Systems with sensors and actuators delivering discrete binary signals that can be modeled as Discrete Event Systems. Even though there are numerous diagnosis methods, none of them can meet all the criteria of implementing an efficient diagnosis system (such as an intelligent solution, an average effort, a reasonable cost, an online diagnosis, fewer false alarms, etc.). In addition, these techniques require either a correct, robust, and representative model of the system or relevant data or experts' knowledge that require continuous updates. In this paper, we propose a Machine Learning-based approach of a diagnostic system. It is considered as a multi-class classifier that predicts the plant state: normal or faulty and what fault that has arisen in the case of failing behavior.

## 1. INTRODUCTION

In order to ensure the safe operation of goods and equipment, the diagnostic task consists of detecting, isolating, and identifying, as accurately and as soon as possible, the slightest failure or deviation from the nominal machine behavior. The context of this work is the diagnosis of Automated Production Systems (APS). In the context of "the industry of the future", production systems need to be more flexible and resilient while becoming more complex. Performance requirements (production, quality, safety) lead manufacturers to avoid stopping their production tool due to breakdowns. The systems we are interested in in this paper are Discrete Event Systems (DES). The classical DES diagnostic approaches in the literature are mainly based on:
1) Offline studies of the diagnosability of a system (ability to diagnose a fault with certainty in a finite time),
2) Online system observer models (diagnosers) to be integrated into the control process.
Although these "diagnoser" approaches are well known by the community, a huge amount of expertise is required to obtain high-performance models of the system. Furthermore, these approaches are quickly exposed to the problem of the explosion of the state space to build the diagnoser of complex systems.

In this paper, we present a new approach based on Machine Learning (ML) techniques using data from normal and abnormal behaviors of a plant. Abnormal behaviors come from a Digital Twin (DT), one of the future industry's tools. The concept of DT (Kritzinger et al., 2018), (Tao et al., 2019) consists in digitizing a factory and reproducing its behavior. Most industrial solutions allow matching a desired behavior of the machine to make virtual commissioning. Here, we look at using it to inject failures into the digitized system to enrich its learning. In (Saddem et al 2022), we have proposed an ML-based approach and recurrent neural networks (RNN) with short-term and long-term memory (LSTM) (Hochreiter, 1997) model to predict the future input/output vector of an APS. In this paper, we have improved this approach. The data acquisition method, the training and validation algorithm, the test, and the architecture of the RNN are different. The diagnostic system is considered as a multi-class classifier that predicts the plant state: normal or faulty and diagnoses what fault has happened in case of fault.

The remainder of this paper is organized as follows: Section 2 presents a brief overview of the state of the art. Section 3 introduces our proposed method. In section 4 we describe an example of an APS on which we will rely to illustrate our approach and we present the results. Finally, we conclude the paper with some prospects in section 5.

## 2. STATE OF THE ART OF DIAGNOSTIC APPROACHES

In this study, we are interested in APS fault diagnosis. The literature proposes different approaches dealing with this problem (Ghosh et al, 2020) and distinguishes three classes according to the dynamics of the APS: the class of continuous systems, the class of DES and the class of hybrid dynamic systems (HDS). In this paper, we focus on the diagnosis of APS's with sensors and actuators delivering logical signals, which fall under the DES. The diagnostic approaches for this class of systems can be seen according to whether the diagnosis is performed online or offline (Sampath et al., 1995), (Boussif and Ghazel, 2021), whether the model is specified (by automaton or by Petri net) or not (Basile, 2014), whether the diagnostic decision-making structure is centralized, decentralized, or distributed, whether faults are represented and recognized, and so on. In general, diagnostic approaches are classified into three main families: model-based approaches, knowledge-based approaches, or data-based approaches.

Model-based approaches (Sampath et al., 1995), (Zaytoon and Lafortune, 2013), are generally used when there is sufficient knowledge of the internal functioning of the system. They are efficient and able to validate the consistency and completeness of the faults to be diagnosed. However, to work properly, these approaches require accurate and deep analytical models of the domain and the major difficulty is the high cost of implementing the models (Saddem and Philippot, 2014), (De Souza et al, 2020), (Moreira and Lesage, 2019). Indeed, the temporal complexity of implementing most models is exponential.

Knowledge-based approaches have a high diagnostic capacity thanks to symptoms of faults they model. However, its major limit lies in the formalization of the expert's knowledge and its updates (Dousson et al, 2008), (Subias et al., 2014).

Data-based approaches (Venkatasubramanian et al., 2003), (Dou and Zhou, 2016), (Han et al., 2017) do not require knowledge of the internal workings of the system. They do not need an explicit formal model. They use available historical data. From this data, they give predictions. These approaches learn from each experience to improve their performance. They rely on ML techniques to achieve their objectives. However, they require a data preparation step to extract the most relevant data that will be formatted according to the ML technique to be used.

In this paper, we are interested in the diagnosis of DES using the data-based approach.

## 3. PROPOSED APPROACH

### 3.1 Automated Production System

An APS system consists of three parts: the operative part (OP), the control part (CP), and the Human Machine Interface (HMI) (Figure 1).

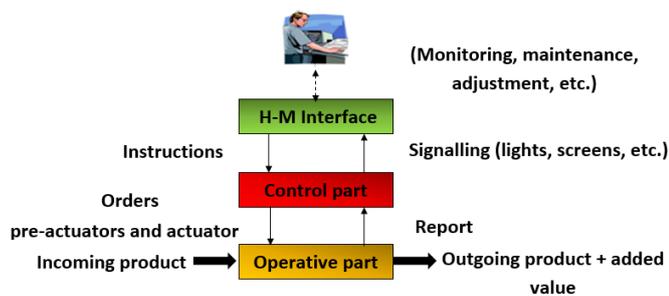

Figure 1: Structure of an APS

OP represents all material resources that physically operate on the system. CP is the set of information processing and acquisition means that ensure the piloting and the control of the process. There are two types of information exchanges between CP and OP i) CP sends orders to the actuators and pre-actuators of OP to obtain the desired effects ii) OP sends reports (sensor values) to CP. HMI allows communication between the CP and the human operator. The human operator gives instructions via HMI and receives various signals from CP such as light indicators, sound indicators, messages displayed on the screens, etc.

Most APS that have sensors and actuators delivering binary signals are controlled by PLC that perform three successive operations: that perform three successive operations: (a) Reading the inputs, which consist of the recording of the states of sensors. (b) Executing the program. (c) Updating the outputs (actuators). These operations are cyclical, i.e. one cycle after the other. The diagnosis, therefore, consists in cyclically reading the sensor's values and the CP's orders and analyzing them to detect and isolate faults.

### 3.2 Proposal

This paper we proposes a new solution for the online diagnosis of APSs that have discrete dynamics. Our solution is based on methods from the field of artificial intelligence (AI). Thus, we use AI techniques to diagnose on line the occurrence of faults of an APS. The development and deployment of ML models involve a series of steps:

i. The definition of the problem, which consists of understanding the problem to be solved, determining the objectives (prediction, clustering, etc.), defining the criteria for success and the constraints to be respected.
ii. Data acquisition, which consists of identifying and collecting data required to support the problem. These data can come from several sources and can be structured (such as database records, trees, graphs…) or unstructured (such as images, texts, voices…)
iii. Data preparation, which consists of formatting the data according to the ML algorithm to be used. It includes transformation, normalization, cleaning, and selection of training data.
iv. The training and validation of the algorithm. This requires dividing the available data into three parts: training data, validation data, and test data. We use cross-validation (CV for short). The training set is split into k smaller sets. First, the model is trained using k−1 of the folds as training data. Then, the resulting model is validated on the Kth fold. The test data is used for testing.
v. The test consists in evaluating the performance of the algorithm.
vi. The deployment of the algorithm.

### 3.3 Understanding the problem

For this step, a study and an analysis of the system are necessary: definition of the list of the APS's components and their operating specifications. In this work, we are interested in the online diagnosis of APS with sensors and actuators delivering binary signals. Four faults are possible for each component: stuck to 0; stuck to 1; an unexpected move from 0 to 1 and an unexpected move from 1 to 0. The monitored APS can be normal, failed, or uncertain. Uncertain state means that the system may be normal or faulty: there are not enough discriminating observations to decide its state. The objective is therefore to return the online status of the plant. If a fault occurs in a component of the plant, the diagnostic may return this fault. Therefore, on needs to have a list of the components of the plant to fix the number and the name of each fault that may occur. A specification of the APS operation allows us to establish a control program for the plant. We assume that this program does not contain any faults, i.e. if CP sends an order to OP, and then OP receives this order correctly.

*3.4 Data Acquisition*

For this step, we have developed an application in JavaScript with the Node-RED software, which allows recovering the data from a PLC. The objective is to save all input and output data history capturing the evolutions of sensors, actuators, and control program variables. These changes are displayed on the screen in the form of charts with periods of activity. The result of this step is an Excel file, which contains the values changes of the monitored variables. Each line represents the name of the variable, the date of occurrence of the change, starting time, and its value.

The developed application allows also the calculation of statistics of the variables. We used this last option to generate symptoms. As an example, if an event B is expected after the occurrence of event A and it does not happen, a symptom is automatically generated.

Our approach starts by collecting data from the APS in normal mode. The use a DT instead of a real APS enables the designer to inject faults on sensors and actuators without any material damage. Several scenarios are possible: for each possible scenario, one collects data from sensors and actuators. It is important to note that the simulated faults of this study are components stuck-off to 1 or to 0.

*3.5 Data Preparation*

This step transforms the Excel files obtained in step 2 into a file containing rows as shown in Figure 2Figure 2. Thus, depending on the architecture used in the training step and depending on the number of past steps to be given to the neural network, data are formatted as labeled values. As an example, the model is fed with the last N past timed input/output vectors in order to predict the state of the system: normal or the faulty state concerned. N is a positive integer that is greater or equal to one.

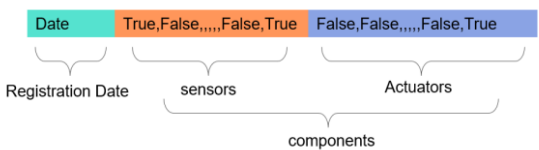

Figure 2: Representation of a timed input/output vector

*3.6 Training of the machine learning algorithm*

The ML model selected by our approach is LSTM RNNs. It is an effective neural model for a wide range of applications involving temporal or sequential data (Karpathy, 2015) such as video analysis, speech recognition (Graves et al, 2013), language modeling, handwriting recognition, or its generation, machine translation, image captioning, etc. Our specific RNN model is depicted by Figure 3.

This work considers eight classes of state noted from $C_0$ to $C_7$. A description of these classes is given in § 4.1.2.b. To obtain a probability distribution at the output of the model, the model uses the softmax function to activate the last layer of the RNN network composed of height neurons (one for each class).

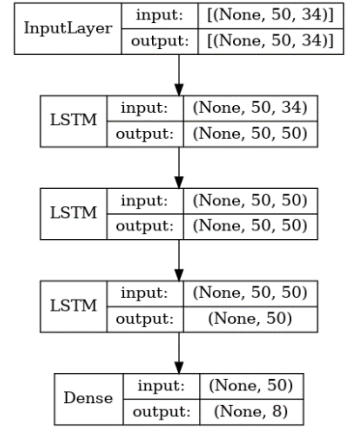

Figure 3 : RNN Model

The output vector of the RNN model returns for each class (possible system state) the probability of belonging to this class. The input vector of the architecture in Figure 3 is generic. The number of timed input/output vectors is given by. N is a hyper-parameter in our model. In Figure 4, the sensors and the actuators are modeled by the letters 'a' to 'z'. T represents the relative time (duration) of the timed input/output vector considering the last change.

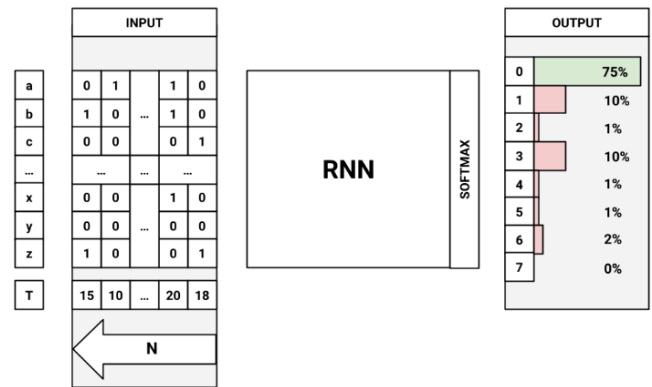

Figure 4: Input-output architecture of the RNN model

The proposed diagnostic system is considered as a multi-class classifier that predicts the plant state. In such classifications, common metrics are being used in the literature to evaluate the classifier's performance. These metrics calculated from a confusion matrix (CM), use the following notations:

- $TP_i$ (True Positive of $C_i$): measures the number of states correctly assigned to $C_i$;

- $TN_i$ (True Negative of $C_i$): measures the number of states correctly recognized that do not belong to $C_i$;

- $FP_i$ (False Positive of $C_i$): measures the number of states that are incorrectly assigned to $C_i$;

- $FN_i$ (False Negative of $C_i$): measures the number of states that are incorrectly recognized that do not belong to $C_i$.

Figure 5 is useful in understanding or visualizing the above-mentioned concepts using a multi-class confusion matrix. In order to confirm the results, cross-validation was used (K-Fold with k = 3). Although the confusion matrix gives us a lot of information about the quality of the classification system, we can calculate other more concise metrics like the average accuracy (AC). It presents the proportion of the correct predictions made by the classifier. It measures the average per-class effectiveness.

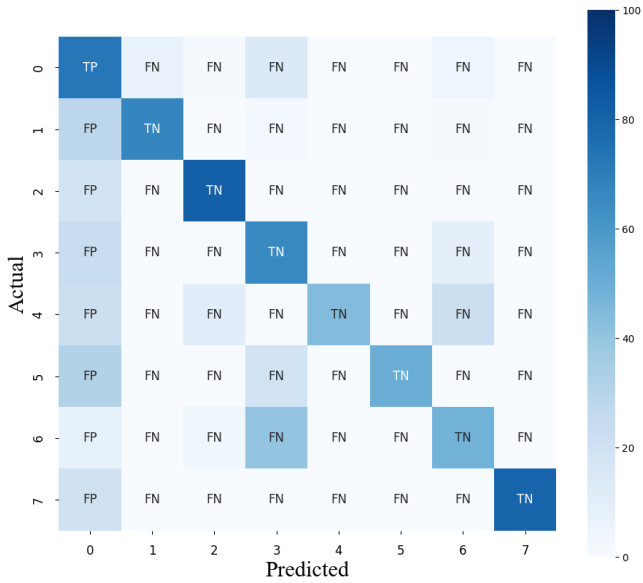

Figure 5 : Illustration of $TP_0, FP_0, TN_0,$ and $FN_0$ for $C_0$ using a multi-class confusion matrix

$$AC = \frac{1}{C} \sum_{i=0}^{C-1} \frac{TP_i + TN_i}{TP_i + TN_i + FP_i + FN_i}$$

With $C$ the number of classes.

We use the following quality metrics:

Precision of $C_i$ called $P_i$ measures the number of cases that the classifier has correctly assigned to $C_i$ divided by the number of cases attributed to this class.

$$P_i = \frac{TP_i}{TP_i + FP_i}$$

Recall of $C_i$ called $RC_i$ measures the proportion of cases correctly assigned to $C_i$ divided by the number of cases belonging to this class.

$$R_i = \frac{TP_i}{TP_i + FN_i}$$

## 4. APPLICATION ON THE REAL SYSTEM

### 4.1 Description of the system

We apply our approach to a part of the Import-Export station of the CellFlex of CReSTIC laboratory in France. Before describing our case study (paragraph 4.1.2), we present the whole CellFlex plant in the following paragraph to give a global view of the considered system.

### 4.1.1 CellFlex

CellFlex is the main component of the CELLFLEX4.0 (https://www.univ-reims.fr/meserp/cellflex-4.0/cellflex-4.0,9503,27026.html) training and research platform at the University of Reims Champagne-Ardenne. The flexible cell, called CellFlex, is a group of eight stations operating together around a central conveyor (Figure 6). These stations simulate the operation of a bottling production line, in the form of a miniaturized factory connected to a network composed of industrial standards.

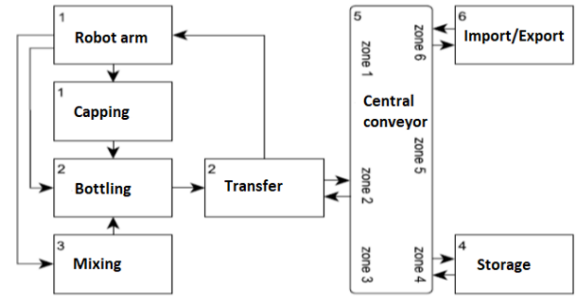

Figure 6: The eight stations of the CellFlex

### 4.1.2 The Import-Export Station

In this work, we focused on the Import-Export station (https://www.univ-reims.fr/meserp/cellflex-4.0/import-export/import-export,15743,27032.html), number 6 in Figure 6. We apply our approach only to the import part, to make easier the comprehension of our model, in this paper.

a) Operation mode

- Import: it consists of introducing a new empty 6-pack on the supply conveyor. When a 6-pack is present at the entrance of the import conveyor, it must be transported to the end of the conveyor, below the vertical cylinder. When an empty pallet is present at the right position on the central conveyor (pallet sensor zone 6), the waiting 6-pack is loaded onto the pallet.
- Export: it consists in taking a full 6-pack from the central conveyor. When a 6-pack is present on the central conveyor, in front of the station, the cylinders must be set in motion in order to take the 6-pack. Then it deposits it on one of the two export lines. The export has priority over the import, in order to avoid a blockage of the system.

b) Description

The import part of the Import-Export station consists of a conveyor allowing the transfer of new 6-packs and two sensors, one at the beginning of the conveyor and the other at the end of the conveyor (Figure 7).

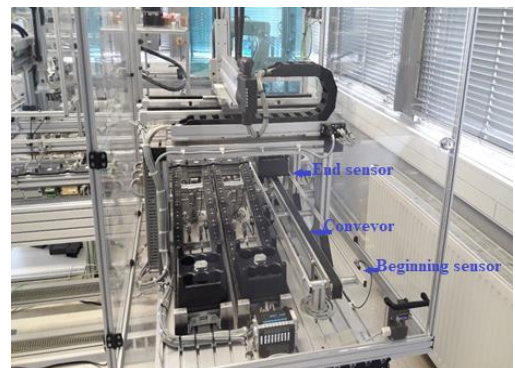

Figure 7 : Import-Export satation

$C_0$ represents the normal behavior of the plant. $C_1$ to $C_6$ represent six faulty behaviors corresponding to stuck to 1 or stuck to 0 of each component (actuator or sensor). $C_7$ represents all other faulty behaviors that can occur on the plant.

4.2 Results

This paragraph resumes the results of the application of our approach on the system described previously. The digital twin of the system is based on the software Emulate 3D. For the data acquisition step (§ 3.4), the system is started in normal mode. The changes in variable values are collected during several cycles. For our system, the cycle duration is of few minutes. To change the system mode, a fault is injected at a moment, using the facilities of the digital twin. Indeed, Emulate 3D allows us to force a sensor or an actuator to a value (true or false). This forcing simulates a stuck to 1 or to 0 of the sensor or the actuator in question. The changes in variable values are collected during several cycles after the injection of the fault). The case study system is composed of 23 sensors and 10 actuators. We simulated stuck to 1 or to 0 of the two sensors and the actuator of the Import part of the studied system.

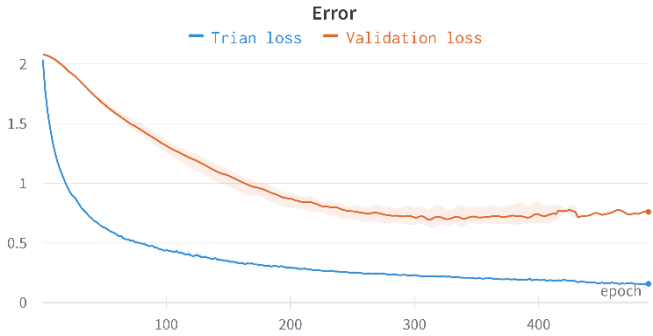

Figure 8: Evolution of the error function

The data preparation transforms the Excel files obtained in the previous step into a file containing lines of the form shown in Figure 2. Next, the date has been formatted as labeled value as described in § 3.5. The vector in Figure 2 is made up of the date and the 33 Boolean values of the system at this date. Then the duration between the previous acquisition step is calculated. Considering the complexity of the system presented in 4.1.2, our approach has estimated that N = 50 is sufficient. An exhaustive search to find the optimal value of N should be performed. The output layer of RNN is activated with the softmax function. For the error function, we use the Categorical Cross-Entropy (CCE) function.

$$CCE = -\sum_{i=0}^{C-1} y_i . \log(\hat{y}_i)$$

Figure 8 and Figure 9 illustrate the obtained results: the evolution of the error function, which converges to zero and the AC that is near 70%. We see that the results are not bad. The AC is difficult to be close to 100%.

In Figure 11, each sub-figure represents the recall (blue curves) and precision (orange curves) on the train (dotted curves) and validation data (solid curves) of each class. For each class, recall and precision results on the validation data tend all to increase overtraining. However, they remain relatively erratic through the three folds for the least present classes (2, 4, 5, and 7). More data representing these types of faults would certainly make it possible to obtain more stable results.

Figure 10 presents the average of normalized multi-class confusion matrix of the three folds.

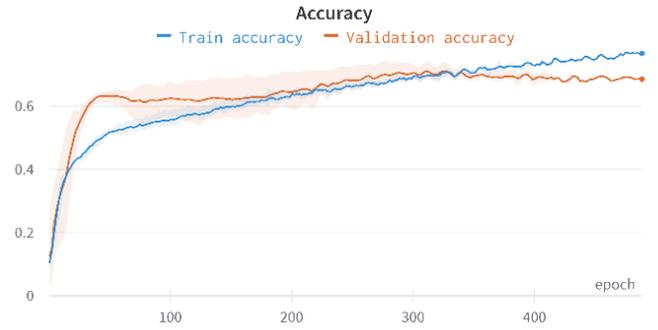

Figure 9 : Evolution of the average accuracy

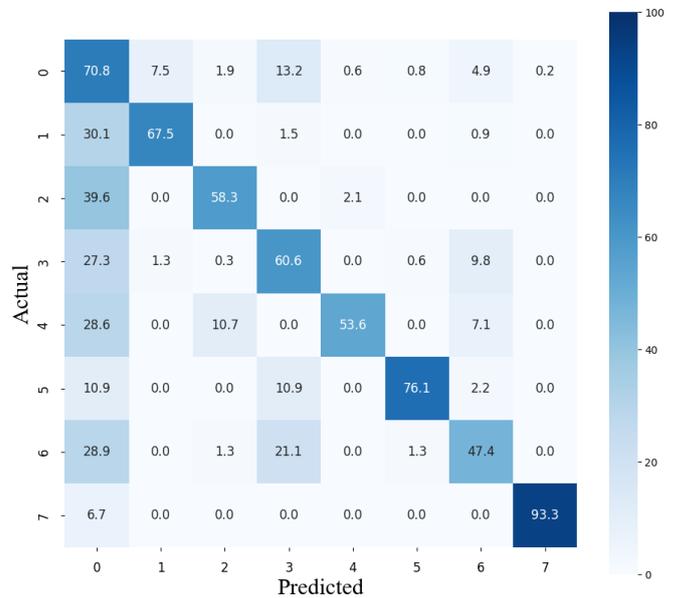

Figure 10 : The average of normalized multi-class confusion matrix of the three folds

5. CONCLUSION

In this paper, we have proposed a new data-based approach for online diagnosis of APS's of DES class. Data acquisition of normal and abnormal behaviors is carried out through a DT using new software developed by our team. The data preparation consists of the transformation of files from this software into vectors for the proposed RNN model. The results of the application of the proposed method on the Import-Export system of CellFlex show the contribution and the interest of this method.

Several perspectives are possible. An exhaustive search to find the optimal value of the hyper-parameters (number of hidden layers, number of neurons on each hidden layer, the size of past observations) could be performed. We have applied the approach to a part of the Import-Export system of the cellFlex which contains 7 other stations, an application on the other stations, and an extension for continuous systems are possible.

A comparison of model-based and knowledge-based diagnosis methods will be carried out in the near future.

## ACKNOWLEDGMENTS

This work is integrated into the project FFCA (Factories of Futur Champagne-Ardenne). The authors would like to thank the region Grand-Est within the project FFCA (CPER PFEXCEL).

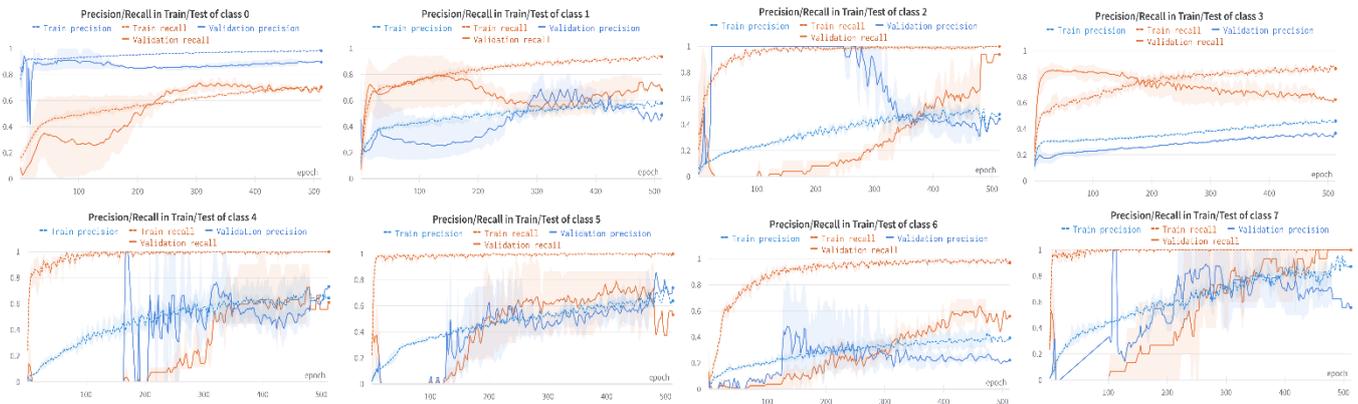

Figure 11: Evolution of the precision and recall of the eight classes of the model